\documentclass{article}
\usepackage{ijcai19}

\usepackage{times}
\usepackage{soul}
\usepackage{url}
\usepackage[hidelinks]{hyperref}
\usepackage{graphicx}
\usepackage{subfig}
\usepackage{amsmath,amssymb,bm,amsthm}
\usepackage{multirow}
\usepackage{xcolor}
\usepackage{booktabs}
\usepackage{cite}
\DeclareMathOperator*{\argmin}{arg\,min}
\title{Integrating Knowledge and Reasoning in Image Understanding}

\author{
Somak Aditya$^1$\and%\footnote{Contact Author}\and
Yezhou Yang$^2$ \And
Chitta Baral$^2$\\
\affiliations
$^1$Bigdata Experience Lab, Adobe Research, India\\ 
$^2$CIDSE, Arizona State University, USA\\
\emails
saditya@adobe.com,
\{yz.yang, chitta\}@asu.edu
}

\begin{document}

\maketitle

\begin{abstract}
  %Incorporating external knowledge sources and higher-level reasoning in image understanding applications is an important avenue. 
  %In the past in computer vision community, there has been a considerable number of successful applications that demonstrated the utility of additional knowledge in image understanding applications; and in the natural language community, the utility of techniques from knowledge representation and reasoning has been demonstrated. Only recently, researchers in computer vision have started exploring ways to integrate knowledge sources and perform higher-level reasoning coupled with deep neural networks.
  
   Deep learning based data-driven approaches have been successfully applied in various image understanding applications ranging from object recognition, semantic segmentation to visual question answering. However, the lack of knowledge integration as well as higher-level reasoning capabilities with the methods still pose a hindrance.  
   In this work, we present a brief survey of a few representative reasoning mechanisms, knowledge integration methods and their corresponding image understanding applications developed by various groups of researchers, approaching the problem from a variety of angles. %This survey also reviews key efforts to acquire large-scale commonsense knowledge bases (about image). 
   Furthermore, we discuss upon key efforts on integrating external knowledge with neural networks. Taking cues from these efforts, we conclude by discussing potential pathways to improve reasoning capabilities.
\end{abstract}

\section{Introduction}

  %Humans beings use a large amount of background and common-sense knowledge to perceive and interact with the environment. This efficient use of knowledge and reasoning with the knowledge makes human beings better in generalization, aids in learning from few examples and helps to interact seamlessly with the surrounding environment while understanding diverse sensory signals (reading, seeing, hearing corresponding respectively to text, vision and speech). 
 %The utility of background knowledge and reasoning has been well known in many applications in artificial intelligence, including natural language and image understanding applications. 
 From the early years of computer vision research, many researchers realized that prior knowledge could help in different tasks ranging from low-level to high-level image understanding. For example, knowledge about the shape of an object can act as a strong prior in segmentation tasks \cite{zheng2015scene}, or knowledge about the most probable action given a subject and the object can aid in action recognition tasks \cite{gupta2009observing,summers2012using}.
 %From the early years of computer vision research, many researchers realized that prior knowledge could help in different tasks ranging from low-level to high-level image understanding. For example, knowledge about the shape of an object can act as a strong prior in segmentation tasks, or knowledge about the most probable action given a subject and the object can aid in action recognition tasks. 
 In this recent era of data-driven techniques, most of this knowledge is hoped to be learned from training data. While this is a promising approach, annotated data can be scarce in certain situations, and many domains have a vast amount of knowledge curated in form of text (structured or unstructured) that can be utilized in such cases. Utilization of background knowledge in data-scarce situations is one of the reasons that necessitate the development of approaches that can utilize such knowledge (from structured or unstructured text) and reason on that knowledge.  Additionally, the lack of reasoning and inference capabilities (such as counterfactual, causal queries) of deep learning systems recently started to resurface in various forums \cite{hudson2018gqa}\footnote{Michael Jordan's views to IEEE Spectrum (\url{bit.ly/2GNejBx}).}.
        \begin{figure}[!htb]
                       \centering \includegraphics[width=0.49\textwidth]{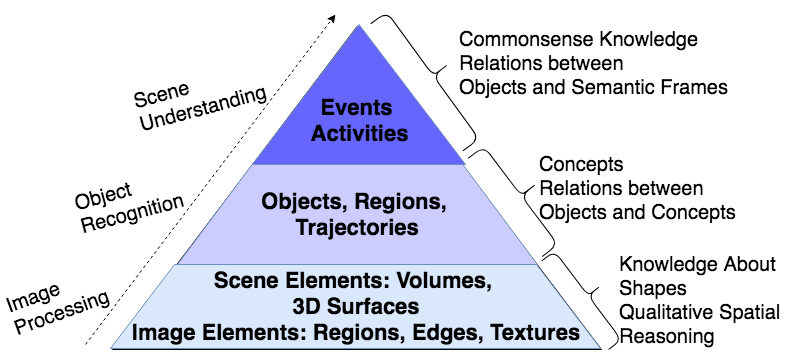} 
                       \caption{The diagram shows the information hierarchy for images and the knowledge associated with each level of information. }
                       \label{fig:hierarchy}
  \end{figure}
 Motivated by these challenges, our goal in this paper is to present a survey of recent works (including a few of our works)  in image understanding where knowledge and reasoning plays an important role. 
 %we first cover the different types of knowledge sources and reasoning mechanisms that have been successfully applied in image understanding applications. 
 While discussing these interesting applications, we introduce corresponding reasoning mechanisms, knowledge sources, and argue the rationale behind their choice.
 %We cover a range of interesting applications and datasets that benefit from knowledge and reasoning mechanisms. 
 \textit{Lastly, we discuss different mechanisms that integrate external knowledge sources directly with deep neural networks.}

   % types of knowledge
   To understand what knowledge is meaningful in images, we can look at the different types of knowledge that relate to different levels of the semantic hierarchy induced by a natural image. Natural images are compositional. A natural image is composed of objects, and regions. Each object is composed of parts that could be objects themselves and regions can be composed of semantically meaningful sub-regions. This compositionality induces a natural hierarchy from pixels to objects (and higher level concepts).  We show a diagram representing the information hierarchy induced by a natural image in Figure \ref{fig:hierarchy}.   
   Different types of knowledge might be relevant in the context of low-level information (objects and their parts)  to higher-level semantics (abstract concepts, actions). Essentially, in this survey, we will study how knowledge and reasoning are applicable to these following levels of semantics: i) objects, regions and their attributes, ii) object-object or object-region interactions, relations and actions; iii) high-level commonsense knowledge (about events, activities).  %The categorization of knowledge sources and applications are inspired from Figure \ref{fig:hierarchy}.

   In this work, we revolve the stories around different image applications ranging from object classification to question answering. As each new reasoning engine and knowledge source is encountered, we introduce them in individual separately marked ($\blacktriangleright$) paragraphs. We provide brief critique as to why the chosen mechanisms were appropriate for the corresponding application. We discuss different ways to integrate knowledge in the deep learning era. The final section then summarizes the reasoning mechanisms. We conclude by shedding light on how the research in  high-level reasoning and utilization of commonsense knowledge in Computer Vision can progress.
   
   %In summary, we first describe the different forms of popular reasoning mechanisms used in the community to reason about images; followed by a detailed description of different kinds of knowledge used by various research groups. We discuss the applications where such integration has been successful, \textit{followed by a discussion of different ways of integration of knowledge sources with deep neural networks. We conclude by shedding light on how the research in  high-level understanding and utilization of commonsense knowledge in Computer Vision can progress}. %how the research in high-level understanding and utilization of commonsense knowledge is lacking. This lack of modeling of knowledge and reasoning is the fundamental motivation behind our approaches towards image understanding applications. We use the following sources of knowledge and types of reasoning mechanisms in our applications presented in the rest of the thesis,

   %our approach in this thesis differs from these approaches and the future directions of this research.
   
%   \section{Related Surveys}
   
%   1. \url{https://arxiv.org/pdf/1705.02908.pdf} MACHINE LEARNING WITH WORLD KNOWLEDGE
%   lake 2016  (Building machines that learn and think like people)
%   3. \url{https://arxiv.org/pdf/1612.04318.pdf} 1) Using prior domain knowledge to prepare training examples; 2)
% Using prior knowledge to initiate the hypothesis or hypothesis space; 3) Using prior domain
% knowledge to alter the search objective; and 4) Using Domain Knowledge to Augment
% Search.

   \section{Use of Knowledge in Computer Vision}
 %   \textbf{\textcolor{red}{Yezhou, Please Shorten Section 4}} 
   %In this section, we discuss different types of knowledge categorized according to different semantic levels of information, induced by a natural image. For each such level in the hierarchy, 
   Here we describe  applications that utilized relevant background knowledge beyond annotated data. Applications are categorized according to the levels of hierarchy in Fig.~ \ref{fig:hierarchy}.

   \subsection{Knowledge about Objects, Regions, Actions}
   
   \subsubsection{Image Classification} Various groups of researchers demonstrated the use of knowledge bases (generic and application-specific) in object, scene and action recognition; or to reason about their properties.
   
%   \\\noindent
%   \fbox{\begin{minipage}{0.96\columnwidth}
 \paragraph{$\blacktriangleright$ Markov Logic Network.} MLN (\cite{richardson2006markov}) is a popular framework that uses weighted First Order Logical formulas to encode an undirected grounded  probabilistic graphical model.  Unlike PSL, the MLN is targeted to use the full expressiveness of First Order Logic and induce uncertainty in reasoning by modeling it using a graphical model.  
   Formally, an MLN $L$ is a set of pairs $\langle F,w \rangle$, where $F$ is a first order formula and $w$ is either a real number or a symbol $\alpha$ denoting hard weight. Together with a finite set of constants $C$, a Markov Network $M_{L,C}$ is defined where: i) $M_{L,C}$ contains one binary node for each grounding of each predicate appearing in $L$; ii) $M_{L,C}$ contains one feature for each grounding of each formula $F_i$ in $L$. The feature value is 1 if ground formula is true otherwise 0.
   The probability distribution over possible worlds %specified by the ground Markov Network $M_{L,C}$ 
   is given by:
   
   \begin{equation*}
   P(X=x) = \frac{1}{Z} \exp (\sum_{i} w_i n_i(x)) = \frac{1}{Z} \prod_{i} \phi_i(x_{i}) ^{n_i(x)},
   \end{equation*}
   where $n_i(x)$ is the number of true groundings of the formula $F_i$ in the world $x$. The MLN inference is equivalent to finding the maximum probable world according to the above formula. Weights are learnt using maximum likelihood methods.
% \end{minipage}}

   Authors in \cite{conf/eccv/ZhuFF14} successfully used Markov Logic Network (MLN) in the context of reasoning about object affordances in images. An example of affordance is \textit{fruit is edible}. Authors collect such assertions from textual cues and image sources, and complete the knowledge base using weighted rules in MLN. Example of collected assertions are \textit{basketball is rollable and round, apple is rollable, edible and a fruit, pear is edible and a fruit} etc. %, depicted in Figure \ref{fig:kb_affordance}. 
   The weights of the grounded rules are learnt by maximizing the pseudo-likelihood given the evidence collected from textual and image sources.
   %Traditional weight learning methods (such as maximum likelihood) are used to learn weights corresponding to the ground rules. 
   Few such weighted rules are: 
   \begin{equation*}
   \small
   \begin{aligned}
   0.82~\text{hasVisualAttr}(x, Saddle) &\implies \text{hasAffordance}(x, SitOn). \\
   0.75~\text{hasVisualAttr}(x, Pedal) &\implies \text{hasAffordance}(x, Lift).\\
   \text{isA}(x, Animal) &\land \text{locate}(x, Below).
   \end{aligned}
   \end{equation*}
   This knowledge base encoded in MLN is used to infer the affordance relationships given the detected objects (and their confidence scores) in an image.
       Besides performing well over pure SVM-based methods, authors also observe that the knowledge-based approach is more robust against removal of visual attributes. %A snapshot of the knowledge base is in Figure \ref{fig:kb_affordance}.
      
% \\\noindent
%   \fbox{\begin{minipage}{\columnwidth}
   \paragraph{Appropriateness.} For modeling object affordances, the authors in \cite{conf/eccv/ZhuFF14} faced the following challenges: i) uncertainty to account for noise and practical variability (if a \textit{pedal} is dis-functional, it cannot be \textit{lifted}); ii) expressive beyond IF-THEN horn-clauses, iii) relational knowledge. Encoding relational knowledge and modeling uncertainty warranted the use of a probabilistic logical mechanisms. As Probabilistic Soft Logic (PSL) can not express beyond horn clauses, and Problog solvers are comparatively slow - MLN was a logical choice for this application.
%   \end{minipage}}
   
   Authors in \cite{serafini2016logic} proposes Logic Tensor Network (LTN) that combines logical symbolism and automatic learning capabilities of neural network. For object-type classification and PartOf relation detection on \textsc{Pascal-Part-Dataset}, LTNs with prior knowledge is shown to improve over Fast-RCNN. This work has started off many contributions in the area of  neuro-symbolic reasoning such as DeepProbLog \cite{manhaeve2018deepproblog}, end-to-end neural networks using prolog \cite{NIPS2017_6969}. 
   
% \\\noindent
%   \fbox{\begin{minipage}{0.96\columnwidth} 
   \paragraph{$\blacktriangleright$ Logic Tensor Network.} In LTN, soft logic is used to represent each concept as a predicate, for example $apple(x)$ represents \textit{apple}. The first-order formula $\forall x\text{ } apple(x) \land red(x) \rightarrow sweet(x)$ represents that \textit{all red apples are sweet}. Here, the truth values of each ground predicates are between 0 to 1, and truth values of conjunctive or disjunctive formulas are computed using combinations functions such as Lukasiewicz's T-norm. To combine this idea of soft logic with end-to-end learning, each concept or predicate is represented by a neural network and objects are represented by points in a vector space. The neural network for ``apple'' takes a point in the feature space and outputs its confidence about the input being a member of the ``apple'' concept. The weights in the neural networks are optimized to abide by rules such as $\forall x\text{ } apple(x) \land red(x) \rightarrow sweet(x)$. These symbolic rules are added as constraints in the final optimization function. 
    % \end{minipage}}
   
   %Frequency or co-occurrence statistics based language models has a long and popular history of usage by the vision and language community, mainly in caption generation applications. However, prior generic commonsense knowledge encoded in large semi-curated knowledge graphs such as ConceptNet can also help language modeling. 
   The authors in \cite{le2013exploiting} uses the commonsense knowledge encoded in ConceptNet to enhance the language model and apply this knowledge to two recognition scenarios: action recognition and object prediction. The authors also carried out a detailed study of how different language models (window-based model topic model, distributional memory) are compatible with the knowledge represented in images. For action recognition, authors detect the human, the object and scenes from static images, and then predict the most likely verb using the language model. They use object-scene, verb-scene and verb-object dependencies learnt from the language models to predict the final action in the scene. %In short, they estimate $P(V|O), P(V|S), P(O|S)$ and $P(O|O)$ from each language model. For the human action
   %recognition scenario, a list of 19 objects, 15 scenes and around 5 thousand verbs are usd for computing $P(V |O), P(O|S), P(V |S)$. 
   Examples of relations extracted from ConceptNet are: \texttt{Oil-Located near-Car, Horse-Related to-Zebra}. The conditional probabilities are computed using the frequency counts of these relations. %For example to predict the probability $P(o_i|s_j)$, authors use:
   %$
   %P(o_i|s_j) = \frac{freq(\langle o_i, rel, s_j\rangle)}{\sum_{o_m \in O} freq(\langle o_m, rel, s_j\rangle)}
   %$. 
   To jointly predict the action i.e. $\langle subject, verb, object \rangle$ triplet the from object, the scene probability and the conditional probabilities from language model, an energy based model is used that jointly reasons on the image (observed variable), object, verb and the scene. %($P(o_j|I), $ $P(s_k|I), P(o_j|s_k)$)
%   \\\noindent
%   \fbox{\begin{minipage}{0.97\columnwidth}
   \paragraph{$\blacktriangleright$ ConceptNet.} Some well-known large-scale commonsense knowledge bases about the natural domain are ConceptNet, (\cite{havasi2007conceptnet}), WordNet (\cite{Miller:1995:WLD:219717.219748}), YAGO (\cite{Suchanek:2007:YCS:1242572.1242667}) and Cyc (\cite{Lenat:1995:CLI:219717.219745}). 
   Among these, ConceptNet is a semi-curated multilingual
Knowledge Graph, that encodes commonsense knowledge about the world and is built primarily to assist systems that attempts to understand natural language text.  The nodes (called concepts) in the graph are words or short phrases written in natural language. The nodes
are connected by edges which are labeled with meaningful relations, such as $\langle reptile, IsA, animal \rangle$, $\langle reptile, HasProperty, cold blood \rangle$. Each edge has an associated confidence score. Being semi-curated, ConceptNet has the advantage of having a large coverage yet less noise.
%   \end{minipage}}
   
%   \subsection{Knowledge about Actions and Activities}
   
%   Authors in \cite{DBLP:conf/aaai/ElhoseinyCCPE17} defines facts (information) of different complexity with respect to images. They define first order facts as objects ($\langle boy \rangle$), second order facts as attributes and actions ($\langle boy,tall\rangle, \langle boy,playing\rangle$) and third order facts as interactions between objects ($\langle boy,riding, horse\rangle$). In this sub-section, we focus on knowledge and reasoning employed to reason about relations and actions that connect two or multiple objects (the third or higher order facts).
    %   \begin{figure}[htpb]
    %      \centering
    %      \includegraphics[width=0.48\textwidth]{survey_figures/tea_prep.png} 
    %       \caption{(Image from \cite{meditskos2014knowledge}) The Relevant Temporally-dependent Observations for the High-level Activity ``Making and drinking Tea''. }
    %      \label{fig:tea_prep}
    %     \end{figure} 
  
  %\subsubsection{Activity Recognition Using Knowledge}

% \\\noindent
%     \fbox{\begin{minipage}{0.97\columnwidth}
    \paragraph{$\blacktriangleright$ Probabilistic Soft Logic.}  Similar to MLN, PSL (\cite{bach2017hinge}) uses a set of weighted First Order Logical rules of the form $w_j: \lor_{i \in I_j^{+}} y_i \leftarrow \land_{i \in I_j^{-}} y_i$, where each $y_i$ and its negation is a literal. The set of grounded rules is used to declare a Markov Random Field, where the confidence scores of the literal is treated a continuous valued random variable. Specifically, a PSL rule-base is used declare Hinge-Loss MRF, which is defined as follows:
      %\begin{definition}
   Let $\bm{y}$ and $\bm{x}$ be two vectors of $n$ and ${n'}$ random variables respectively, over $D =[0,1]^{n+{n'}}$. Let $\tilde{D} \subset D$, which satisfies a set of inequality constraints over the random variables.
   %For $(\bm{y},\bm{x}) \in \tilde{D}$, and given a vector of non-negative weights $\bm{w}$ and real-valued potential functions $\bm{\phi}(.)$, the hinge-loss energy function is defined as: $f_{\bm{w}}(\bm{y},\bm{x}) = \bm{w}^{T}\bm{\phi}(\bm{y},\bm{x}) = \sum_{j=1}^{m} w_j \phi_j(\bm{y},\bm{x})$.
    %   \begin{equation*}
    %   f_{\bm{w}}(\bm{y},\bm{x}) = \sum_{j=1}^{m} w_j \phi_j(\bm{y},\bm{x})
    %   \end{equation*}
   A \textit{Hinge-Loss MRF} $\mathbb{P}$ is a probability density over $D$, defined as: if $(\bm{y},\bm{x}) \notin \tilde{D}$, then $\mathbb{P}(\bm{y}|\bm{x})=0$; if $(\bm{y},\bm{x}) \in \tilde{D}$, then:
      $
      \mathbb{P}(\bm{y}|\bm{x}) = \frac{1}{Z(\bm{w})}  exp(-f_{\bm{w}}(\bm{y},\bm{x}))
      $.   Using PSL, the hinge-loss energy function $f_{\bm{w}}(\bm{y})$ is defined as:
      \begin{equation*}
      \sum\limits_{C_j \in \bm{C}} w_j\text{ }max\big\{ 1- \sum_{i \in I_j^{+}} V(y_i) - \sum_{i \in I_j^{-}} (1- V(y_i)),0\big\},
      \end{equation*} 
      where $max\big\{ 1- \sum_{i \in I_j^{+}} V(y_i) - \sum_{i \in I_j^{-}} (1- V(y_i)),0\big\}$ is distance to satisfaction of a grounded rule $C_j$.
      MPE inference in HL-MRFs is equivalent to finding a feasible minimizer for the convex energy function, and in PSL it is equivalent to $\argmin_{\bm{y}\in [0,1]^n} f_{\bm{w}}(\bm{y})$. 
To learn the parameters $\bm{w}$ of an HL-MRF from the training data, maximum likelihood estimation (and its alternatives) is used. %Alternatives such as maximum pseudo-likelihood is used for fast learning.
    % \end{minipage}} 
    
  \subsubsection{Actions and Activities} 
  Authors in \cite{london2013collective} uses PSL to detect collective activities (i.e. activity of a group of people) such as  \textit{crossing, queuing, waiting} and  \textit{dancing} in videos.  This task is treated as a high-level vision task, whereby detection modules are employed to extract information from the frames of the videos and such information (class labels and confidence scores of predicates) is input to the joint PSL model for reasoning. To obtain frame-level and person-level activity beliefs, human figures are represented using HOG features and Action Context (AC) descriptors. %(\cite{lan2010retrieving}).  
   Then an SVM is used to obtain activity beliefs using these AC descriptors. Next, a collection of PSL rules is used to declare the ground HL-MRF to perform global reasoning about collective activities:
   {
   \begin{equation*}
   \begin{aligned}
   \textsc{LOCAL}(B, a) &\implies \textsc{DOING}(B, a). \\
   \textsc{FRAME(B, F)} \land \textsc{FRAMELBL(F, a}) &\implies \textsc{DOING(B, a)}.
   \end{aligned}
   \end{equation*}
   }
   The intuitions behind the two rules are: Rule R1 corresponds to beliefs about local predictions using HOG features, and R2 expresses
   the belief that if many actors in the current frame are doing
   a particular action, then perhaps everyone is doing that
   action. To implement this, a \textit{FrameLbl} predicate
   for each frame is computed by accumulating and normalizing the \textit{Local} activity beliefs for all actors in the frame. Similarly, there are other rules that captures the intuition about these activities. Using PSL inference, final confidence scores are obtained for each collective activity, and authors observe that using PSL over baseline HOG features achieves significant increase ($10\%$) in accuracy.
  
%   \\\noindent
%   \fbox{\begin{minipage}{\columnwidth}
   \paragraph{Appropriateness.} To model group activities, the authors in \cite{london2013collective} faced the following challenges: i) uncertainty to account for noise in real-world data and noisy predictions form machine learning models; ii) a fast scalable mechanism to predict activity classes. In the presence of large annotated data, deep learning models are the de facto standards for any kind of classification. However, smaller datasets and the requirement of interpretability warranted a logical reasoning language. Requirement of robustness to noise and scalability warranted the use of PSL, as its underlying optimization problem is convex and its found to be faster. % than other mechanisms.
%   \end{minipage}}

  \paragraph{Infrequently Used Logical Languages.} The reasoning mechanisms discussed in this survey are chosen based on the following considerations: i) plausible inference, ii) learning capability, iii) expressiveness, and iv) speed of inference \cite{Davis:2015:CRC:2817191.2701413}. Several other logical languages have factored in useful aspects such as uncertainty, spatio-temporal reasoning etc. Qualitative Spatial Reasoning \cite{Randell92aspatial} languages and description logic \cite{baader2003description} are noteworthy among them for image understanding. A popular representation formalism in QSR is Region Connection Calculus (RCC) introduced in  \cite{Randell92aspatial}. The RCC-8 is a subset of the original
RCC. It consists of the eight base relations: disconnected (DC), externally connected
(EC), partial overlap (PO), equal (EQ), tangential proper-part (TPP), non-tangential
proper-part (NTTP), tangential proper-part inverse (TPP$^{-1}$), and non-tangential properpart inverse (NTPP$^{-1}$). Extensions of RCC-8 is used to successfully reason about visuo-spatial dynamics, and (eye-tracking based) visual perception of the moving image in cognitive film studies (\cite{suchan2016geometry}). \textbf{Description Logics} (\cite{baader2003description}) model relationships between entities in a particular domain. In DL, three kind of entities are considered, concepts, roles and individual names. Concepts represent classes (or sets) of individuals, roles represent binary relations between individuals and individual names represent individuals (instances of the class) in the domain. Fuzzy DLs extend the model theoretic semantics of classical DLs  to fuzzy sets. In \cite{dasiopoulou2009applying}, Fuzzy DL is used to reason and check consistency on object-level and scene-level classification systems. 

     \begin{figure*}[!htpb]
              \centering
              \subfloat{\includegraphics[width =0.3\textwidth, height=0.2\textheight]{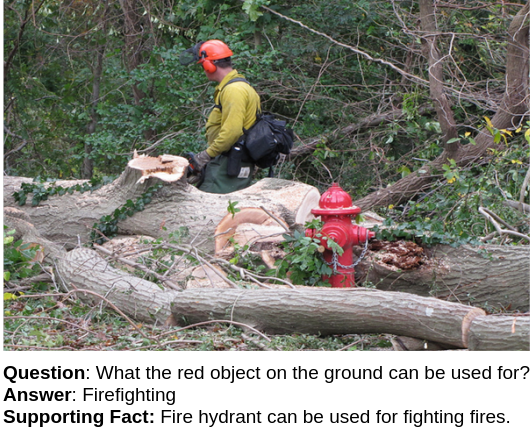}}
              \subfloat{\includegraphics[width =0.3\textwidth, height=0.2\textheight]{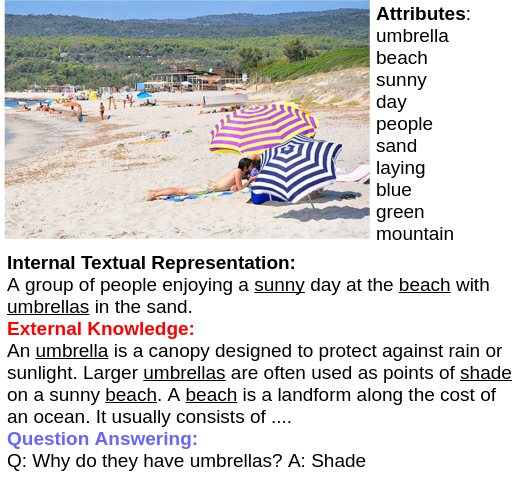}}
              \subfloat{\includegraphics[width=0.35\textwidth, height=0.2\textheight]{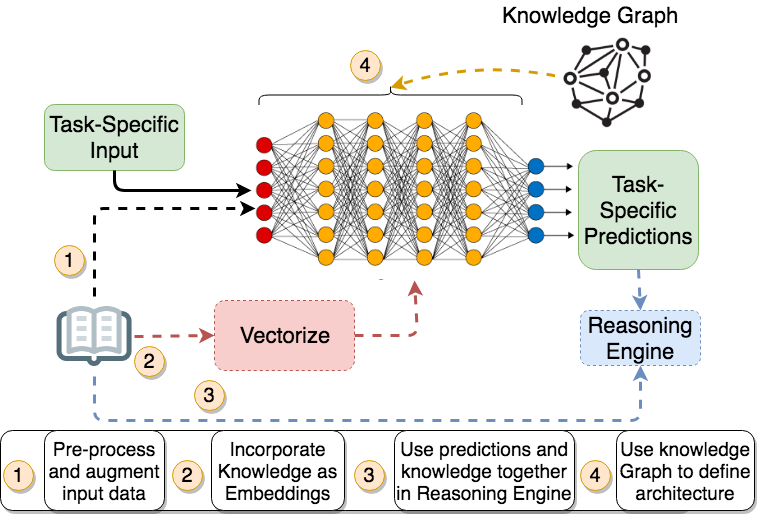}}
               \caption{(a) Example of questions that require explicit external knowledge \cite{wang2017fvqa}, (b) Example where knowledge helps \cite{wu2016ask}. (c) Ways to integrate background knowledge: i) Pre-process knowledge and augment input \cite{aditya2018spatial}; ii) Incorporate knowledge as embeddings \cite{wu2017visual}; iii) Post-processing using explicit reasoning mechanism \cite{aditya2018vqa}; iv) Using knowledge graph to influence NN architecture \cite{marino2017more}.}
              \label{fig:fvqa}
             \end{figure*}

  %Researchers in robotics  %\cite{ye2017can} 
  %and computer vision \cite{miyanishi2018generating} have used scene-related ontology and domain knowledge to detect functionality and composite activities. The presence of application-specific heuristics makes the applicability of the methods limited.
  
  %Authors in \cite{meditskos2014knowledge} demonstrated the use of an RDF dataset of primitive observation and traditional Description Logic reasoning to recognize higher-level activities in limited setting such as tea preparation. Given a set of video clips in a clinical setting, the authors propose to extract objects which denote low-level observations such as objects, locations postures and are connected to RDF instances in the available RDF ontology. Given a set of pre-determined domain descriptors, the authors propose to identify meaningful contexts in $O$ to detect higher-level activities. For example, context descriptors such as \textit{Drinking, TeaCup, Sitting, Table, TableZone} are used to detect the higher-level activity \textit{Making and Drinking Tea}.  The low-level relevant observations are summarized in Figure \ref{fig:tea_prep}. The authors have demonstrated their result on 10 daily activities including ``Prepare hot tea'', ``Make a phone call'', ``Watch TV'' etc. 

   %Part-Whole Relations Mining Tandon 
   %Knowledge-Driven Activity Recognition and Segmentation Using Context Connections (Meditskps et. al.)

   \subsection{High-level Common-sense Knowledge}
   
   Several researchers employed commonsense knowledge to enrich high-level understanding tasks such as visual question answering, zero-shot object detection, relationship detection. Answering questions beyond scene information, detecting objects in data-scarce or partial observable situation are natural candidates for employing reasoning with knowledge.
   
%   \\\noindent
%   \fbox{\begin{minipage}{0.97\columnwidth}
   \paragraph{$\blacktriangleright$ Graph-Gated Neural Network (GGNN).} Given a graph of $N$ nodes, at each time-step GGNN produces some output for each node  $o_1,o_2,...o_N$ or global output $o_G$. The propagation model is similar to an LSTM.  For each node $v$ in the graph, there is a corresponding hidden state $h_v^{(t)}$ at every step $t$. At $t=0$, they are initialized with initial state $x_v$, for example for a graph of object-object interactions, it is initialized as one bit activation representing whether an object is present in an image. 	 Next, the structure of the graph is used (encoded in adjacency matrix $A$) along with the gated update module to update hidden states. The following equations summarize the update for each timesteps:
   \begin{equation*}
   \begin{aligned}
   h_v^{(1)} &= [x_v^{T},0]^T. \\
   a_v^{(t)} &= A_v^{T}[h_1^{(t-1)},\ldots,h_N^{(t-1)}]^T + b.\\
   z_v^{t} &= \sigma(W^za_v^{(t)} + U^zh_v^{(t-1)}). \\
   r_v^{t} &= \sigma(W^ra_v^{(t)} + U^rh_v^{(t-1)}). \\
\tilde{h_v^{t}} &= tanh(Wa_v^{(t)} + U(r_v^{t} \odot h_v^{(t-1)})). \\
h_v^{(t)} &= (1-z_v^{t}) \odot h_v^{(t-1)} + z_v^t \odot \tilde{h_v^{t}},
   \end{aligned}
   \end{equation*}
   where $h_v^{t}$ is the hidden state of node $v$ at timestep $t$ and $x_v$ is the initial specific annotation. After $T$ timesteps, node-level outputs can be computed as: $o_v = g(h_v^{T}, x_v)$.
   
%   \end{minipage}}
   
   \subsubsection{Image Classification}
   Authors in \cite{marino2017more} employed the structured prior knowledge of similar objects and their relationships to improve end-to-end object classification task.  Authors utilize the notion that humans can understand a definition of an object written in text and leverage such understanding to identify objects in an image. The authors introduce Graph Search Neural Network to utilize a knowledge graph about objects to aid in (zero-shot) object detection. This network uses image features to efficiently annotate the graph, select a relevant subset of the input graph and predict outputs on nodes representing visual concepts. GSNN learns a propagation model which reasons
   about different types of relationships and concepts to produce outputs on the nodes which are then used for image classification. The knowledge graph is created from Visual Genome by considering object-object and object-attribute relationships. %Next we introduce Graph-Gated Neural Network and the change suggested by GSNN.
   
     For GSNN, the authors propose that rather than performing recurrent updates over entire graph like GGNN, only a few initial nodes are chosen and nodes are expanded if they are useful for the final output. For example, initial nodes are chosen based on the confidence from an object detector (using a threshold). Next, the neighbors are added to the active set. After each
   propagation step, for every node in our current graph, authors predict an importance score using the importance network: 
   $i_v^{t} = g_i(h_v, x_v)$. 
   This importance network is also learnt. Based on the score, only top $P$ scoring non-expanded nodes are selected and added to the active set. The structure (nodes and edges) of the GSNN can be initialized according to ConceptNet or other knowledge graphs, thereby directly incorporating external knowledge. As GSNN can be trained in an end-to-end manner, this approach provides distinct advantages over sequential architectures. On Visual Genome multi-label classification, the authors achieve significant accuracy over a VGG-baseline using combined knowledge from visual genome and WordNet. %However, training and initializing GSNN becomes harder as the underlying knowledge graph gets larger.
%   \\\noindent
%   \fbox{\begin{minipage}{\columnwidth}

   \paragraph{Appropriateness.} Authors in \cite{marino2017more} utilizes GSNN purely for knowledge integration purpose, i.e. to enhance an object classifier by using ontological knowledge about objects. An alternative would be use object classifier first and then use PSL or MLN to reason i.e. refine the final output - this method however cannot backpropagate the errors to the classifier. 
%   \end{minipage}}

            % \begin{figure}[htpb]
            %   \centering
            %   \includegraphics[width=0.45\textwidth]{survey_figures/knowledge_vqa1.png}
            %   \caption{(Example From \cite{wu2016ask}) An Example Image, Question and Answer with Usable External Knowledge.}
            %   \label{fig:fvqa}
            %  \end{figure}
             
   \subsubsection{Knowledge in Question-Answering} Authors in \cite{wang2017fvqa,shahMYP19} observed that %that even though the task of visual question answering requires reasoning with external knowledge, 
   popular datasets do not emphasize on questions that require access to external knowledge. The authors \cite{wang2017fvqa} proposed a new dataset named Fact-based VQA (or FVQA) where all questions require access to external (factual or commonsense) knowledge that is absent in the input image and the question. A popular example from their dataset is presented in Figure \ref{fig:fvqa}. The questions are generated using common-sense facts about visual knowledge which is extracted from ConceptNet, DBPedia, WebChild. In the proposed approach, structured predicates are predicted using LSTM from the question. For the question \textit{Which animal in the image is able to climb trees}, the generated query example is $\{?X, ?Y\} = Query("Img1", "CapableOf", "Object")$.  Then a set of object detector, scene detectors and attribute classifiers are used to extract objects, scenes and attributes from the image. This query is fired against the knowledge base of RDF triplets, and the answers are matched against the information extracted from the image. 
   
   %Using external knowledge to answer questions about an image has been only recently popular in the community. 
   Authors in \cite{wu2016ask} use knowledge in web-sources to answer visual questions. %For an image with a
%   \textit{Caption}: A group of people enjoying a sunny day at the beach with umbrellas in the sand.\\
%   \textit{Knowledge}: An umbrella is a canopy designed to protect against train or sunlight. Larger umbrellas are often used as points of shade on a sunny beach. ...\\
%   \textit{Question} Why do they have umbrellas? \textit{A}: Shade
   They propose to use fixed-length vector representations of external textual description paragraphs about objects present in the image in an end-to-end fashion. For example, for an image about a dog, a Multi-label CNN classifier is used to extract top 5 attributes, which are then used to form a SPARQL query against DBPedia to extract the definition paragraph about relevant objects. The Doc2vec representation of this paragraph is then used to initialize the hidden state at the initial time-step of the LSTM that ultimately processes the question-words in their end-to-end question-answering architecture. The example of an image, question and relevant external knowledge is provided in the Fig.~\ref{fig:fvqa}. On Toronto Coco-QA dataset, the authors achieve a sharp $11\%$ increase in accuracy after using knowledge sources.

      %\textcolor{red}{  
      %5. ???Ask Me Anything: VQA Based on Knowledge from External Sources??? Wu et al. https://arxiv.org/pdf/1511.06973.pdf
      %6. \url{http://www.cs.toronto.edu/~fidler/slides/2017/CSC2539/vicol_knowledgebasevision.pdf}
      %7. ???Explicit Knowledge-Based Reasoning for VQA.??? Wang et al. https://arxiv.org/pdf/1511.02570.pdf. 2015}

   \paragraph{Visual Relationship Detection.}
   Knowledge distillation has been effective in integrating external knowledge (rules, additional supervision etc.) in natural language processing applications. 
   Authors in \cite{yu2017_vrd_knowledge_distillation} incorporate subject-object correlations from ConceptNet using knowledge distillation. %A significant work that uses the knowledge distillation framework to distill knowledge in image applications is by authors in \cite{yu2017_vrd_knowledge_distillation}. 
   Authors use linguistic knowledge from ConceptNet to predict conditional probabilities ($P(pred|subj,obj)$) to detect visual relationships from image. \cite{aditya2018spatial} uses knowledge distillation to integrate additional supervision about objects to answer questions.
   
      \paragraph{Knowledge in Image Retrieval.}
   Authors in \cite{de2015applying} observed the semantic gap between high-level natural language query and low-level sensor data (images), and proposed to bridge the gap using rules and knowledge graph such as ConceptNet. They proposed a semantic search engine, where the dependency graph of a query is enhanced using hand-written rules and ConceptNet to match scene elements. The enhanced graph is used rank and retrieve images. % which transforms words in the dependency graph of a query to scene elements using hand-written rules. To avoid exact string matches, ConceptNet is used to expand the query and then 
   %that retrieves images given natural language queries. The user queries are first interpreted using a syntactic dependency parser. The dependency graph is passed to a Semantic Interpretation module, where \textit{hand-written rules} are used to transform elements of the graph to semantic scene elements such as \textit{objects, actions, scenes} and \textit{relations}. This graph is then sent to a Semantic Analysis module that matches the graph nodes against the available image concepts. If exact-match is not found, query is \textit{expanded using ConceptNet} to find a match. The query graph is then used as input to a Retrieval and Ranking module. %We provide an example in Figure \ref{fig:query_exp}.

   \section{Discussion: Knowledge Integration in the Deep Learning Era}
    % \begin{figure}[!htpb]
    % \centering
    %       \includegraphics[width=0.5\textwidth]{survey_figures/knowledge2.png} 
    %         \caption{Ways to integrate background knowledge: i) Pre-process knowledge and augment input \cite{aditya2018spatial}; ii) Incorporate knowledge as embeddings \cite{wu2017visual}; iii) Post-processing using explicit reasoning mechanism \cite{aditya2018vqa}; iv) Using knowledge graph to influence NN architecture.}
    %       \label{fig:knowledge}
    % \end{figure}

    In this era where differentiable neural modules are dominating the state-of-the-art, a natural question arises as to how to integrate external knowledge with deep neural networks. The machine learning community (\cite{wulfmeier2016incorporating}) have explored this in terms of constraining the search space of an optimization (or machine learning) algorithm. Four ways are adopted to use prior domain knowledge:  1) preparing training examples; 2) initiating the hypothesis or hypothesis space; 3) altering the search objective; and 4) augmenting search process. Along similar lines, we discuss four primary ways (shown in Fig.~\ref{fig:fvqa}(c)) to integrate knowledge into deep neural networks: i) pre-process domain knowledge and augment training samples, ii)  vectorize parts of knowledge base and input to intermediate layers, iii) inspire neural network architecture from an underlying knowledge graph,  iv) post-process  and reason with external knowledge. For each type, we provide a few recent works that have shown success along the line of increased accuracy or interpretability.

    Pre-processing knowledge and using it as input to starting or intermediate layers is a popular intuitive way of integrating knowledge. For example, authors in \cite{wu2016ask} have vectorized relevant external textual documents and used the knowledge to answer knowledge-based visual questions. Similarly, other authors %\cite{he2017reinforcement} 
    have used similar techniques in deep reinforcement learning.  However, additional fact-based knowledge can be noisy and may bias the learning procedure. %In the presence of ground-truth annotations and external knowledge, a mechanism is needed which can learn to balance between the two without being biased. 
    The knowledge distillation framework is effective for learning to balance between ground-truth annotations and external knowledge.
    
%             \\\noindent
% \fbox{\begin{minipage}{0.97\columnwidth}
\paragraph{$\blacktriangleright$ Relational Reasoning Layer.}
%   The KR\&R reasoning languages such as Answer Set Programming (ASP), Prolog often use $2$-ary 
%   predicates to describe the current world, such as $color(object_1, red)$, $shape(object_1,sphere)$, $material(object_1,metal)$ etc; and then declare rules that the world should satisfy.  Using these rules, truth values of unknown predicates are obtained, such as $ans(?x)$ etc. For example, for a query ``find metallic spherical objects'', a rule in ASP can be expressed as:
%   $
%   ans(x) \leftarrow object(x) \land shape(x, Sphere) \land material(x,Metal).
%   $
%   As output, we get the truth value of $ans(object_1)$ to be true.
   Authors in \cite{santoro2017simple} defined a relational reasoning layer that can be used as a module in an end-to-end deep neural network and trained using traditional gradient descent optimization methods. This module takes as input a set of objects, learns the relationship between each pair of objects, and infer a joint probability based on these relationships (with or without the context of a condition such as a question). Mathematically, the layer (without the appended condition vector) can be expressed as:
   $
    RN(O) = f_\phi \Big(\sum_{i,j} g_\theta(o_i,o_j)\Big),
   $ 
    where $O$ denote the list of input objects $o_1,\ldots, o_n$. In this work, the relation between a pair of objects  (i.e. $g_\theta$) and the final function over this collection of relationships i.e. $f_\phi$ are modeled using multilayer perceptrons and are learnt using gradient descent. %This model's simplicity and its close resemblance to traditional reasoning mechanisms makes the work attractive to be usable for a wide range of applications in image understanding.
% \end{minipage}} 

    %     \\\noindent
    % \fbox{\begin{minipage}{0.96\columnwidth}
    \paragraph{$\blacktriangleright$ Knowledge Distillation.} is a generic framework (\cite{hinton14071698}) where there are two networks, namely the teacher and  the student network. There are two traditional settings, i) teacher with additional computing layers, ii) teacher with additional knowledge. 
      In the first setting, the teacher network is a much deeper (and/or wider) network with more layers. The teacher is trained using ground-truth supervision where in the last layer \texttt{softmax} is applied with a higher temperature (ensuring smoothness of values, while keeping the relative order). The student network, is a smaller network that aims to compress the knowledge learnt by the teacher network by emulating the teacher's predictions. 
In the second setting popularized in natural language processing and computer vision, the teacher network is a similar-sized network which has access to external knowledge, so that it learns both from ground-truth supervision and the external knowledge. The student network, in turn, learns from ground-truth data and teacher's soft prediction vector. The student network's loss is weighted according to an imitation parameter that signifies how much the student can trust the teacher's predictions over groundtruth. %It has often been observed that rather than learning sequentially (i.e. student learning from a pre-trained teacher), it is often beneficial to learn iteratively where the teacher's loss component includes a loss comparative to the student's predictions as well.
    % \end{minipage}} 

    Often question-answering datasets (CLEVR \cite{johnson2017clevr}, Sort-of-Clver, Visual Genome \cite{krishnavisualgenome}) have additional annotations such as properties, labels and bounding box information of the objects, and the spatial relations among the objects. %These annotations are available for large datasets such as CLEVR, Sort-of-Clevr and Visual Genome. 
    Authors in \cite{aditya2018spatial} utilized this \textit{knowledge} in a framework that also considers the non-availability of such information during inference time. As \textit{preprocessing}, the authors use PSL engine to reason with the structured source of knowledge about objects and their spatial relations, and the question, and construct a pre-processed attention mask for training image-question pairs. For an image-question pair, this attention mask blocks out the objects (and regions) not referred to in the question. As \textit{reasoning mechanism}, authors use the combination of knowledge distillation and relational reasoning \cite{santoro2017simple} which achieves a $13\%$ increase in accuracy over the baseline. Relational Reasoning mechanism is also used for visual reasoning in \cite{santoro2017simple}, VQA in \cite{Cadene_2019_CVPR}, temporal reasoning in \cite{Zhou_2018_ECCV}. %The knowledge distillation paradigm requires two networks, namely the teacher network and the student network. 
    %During training, the teacher learns from the ground-truth answers and the additional knowledge, and the student learns from ground-truth data and teacher's soft predictions.

% \\\noindent
% \fbox{\begin{minipage}{\columnwidth}
\paragraph{Appropriateness.} Authors in \cite{aditya2018spatial,santoro2017simple} used relational reasoning for image question answering. They model relationships as functions between objects and use these functions together to answer questions about an image. Despite several efforts of creating scene graphs \cite{johnsoncvpr2015,aditya2017image}, defining a closed set of complete relationships between objects is nontrivial. Hence modeling the relationships as functions is equally acceptable practice. Modeling the reasoning as a function on these triplets is similar to that of PSL, MLN. But learning objects, relationships and the function together is why the relational reasoning layer is a popular choice. Though, semantics associated with this function ($f_\phi$) is hardly understood, which makes it a cautionary tale for out-of-the-box reasoning alternative.
% \end{minipage}}

Reasoning with outputs from deep neural networks and utilizing structured knowledge predicates is another natural alternative. 
    %Similar to pre-processing, post-processing deep learning outputs using reasoning mechanisms while integrating external knowledge is a natural alternative. 
    As most off-the-shelf reasoning engines suffer because of issues of scalability, and uncertaintly  modeling; authors in \cite{aditya2018answering,aditya2018vqa} develop a PSL engine that achieves fast inference on weighted rules with open-ended structured predicates and applied it to solve image puzzles and visual question-answering.
    %However, the most robust state-of-the-art reasoning mechanisms are limited because of issues of scalability, uncertaintly  modeling and numerical reasoning.  
    %Authors in \cite{aditya2018answering,aditya2018vqa} develop a probabilistic soft logic engine that achieves fast inference on weighted rules with open-ended structured predicates and applied it to solve image puzzles and visual question-answering. %(\cite{aditya2018answering}) and VQA (\cite{aditya2018vqa}), authors develop a probabilistic soft logic engine that understands similarities between predicates that are natural language phrases to reason with the noisy data VQA and image puzzle solving tasks with open-ended phrases as relations.  
    For both tasks, authors use ConceptNet and pre-learnt word2vec embeddings as knowledge sources. In the image puzzle solving, the task is to find a common meaningful concept among multiple images.
    %\footnote{Such as the word "fall" connects images of waterfall, rainfall, fall season and statue-falling.}. 
    Authors use an off-the-shelf image classifier algorithm to predict concepts (or objects) present in each image; followed by a set of simple propositional rules in PSL such as $w_{ij}: s_i \rightarrow t_j$, where $s_i$ is a predicted class, $t_j$ is a target concept from ConceptNet vocabulary. The weight of the rule $w_{ij}$ is computed by considering the (ConceptNet-based) similarity of the predicted class ($s_i$) and the target concept ($t_j$), and the popularity of the predicted class in ConceptNet. %Note that, the ConceptNet-based similarity embodies the strength of all the shortest paths connecting the two concepts. 
    Reasoning with the rules of this form, authors predict the most probable set of targets from a larger vocabulary given class-predictions (and their scores). Using a similar rule-base, authors then jointly predict the most probable common targets for all images, which provides the final ranking of concepts. Additionally for the VQA task in \cite{aditya2018vqa}, authors first obtain textual information from images using dense captioning methods. Then they parse the question and the captions using a rule-based semantic parser to create semantic graphs; and use these two knowledge structures in the reasoning engine to answer the question. To understand open-ended relations, ConceptNet and word2vec is used. The solution is shown to increase accuracy over state-of-the-art for "what" and "which" questions. The reasoning engine can be used to predict structured predicates as evidence along-with the answer, aiding in increased interpretability of the system.
    
    Another important contribution is to utilize the nodes and connections of publicly available knowledge-graphs (such as ConceptNet) to build a Neural Network. As explained before, authors in \cite{marino2017more} have used this technique for a more robust image classification. %As the original techniques in GGNN suffers due to the vastness of public knowledge graphs, 
    Authors have improved upon GGNN to propose Graph-search Neural Network that lazily expands the nodes when they are encountered during training. However, the approach is only shown to works on sub-graphs of a large knowledge graph, and does not have explicit consideration for handling incompleteness in the graph.

   \section{Summary and Future Works}
  %{\textcolor{red}{Yezhou and Somak}} 
  
  In this paper we have discussed several reasoning mechanisms such as PSL, MLN, LTN, relational reasoning layers and their use in various image understanding applications. Here we give a quick summary of our assessment of these reasoning mechanisms. Early researchers in AI realized the importance of knowledge representation and reasoning and also realized that classical logics (such as first-order logic) may not be suitable for reasoning in where one may have to retract an earlier conclusion, when presented with new knowledge. This led to the development of various non-monotonic logics such as Answer Set Programming (ASP). Recent extensions of it, such as P-log, Problog \cite{DeRaedt:2007:PPP:1625275.1625673} and LP-MLN \cite{lee2017computing} allow expression of probabilistic uncertainty, weights and contradictory information to various degrees. There are also recent works that extend Inductive Logic Programming techniques to learn ASP rules and also to learn weights. However, like MLN, which can be thought of as extension of first order logic, ASP has high computational complexity, even when the set of ground atoms are finite. PSL, uses a restricted syntax for its rules (thus is less expressive than the others), does not have non-monotonic features, requires its ground atoms to have continuous truth values and uses characterizations of logical operations so that its space of interpretations with nonzero density forms a convex polytope. This makes inference in PSL a convex optimization problem in continuous space, increasing efficiency of inference. Many description logics are decidable fragments of first-order logic (FOL) with focus on reasoning concepts, roles and individuals, and their relationships. 
  %In the probabilistic logics, it is hard to learn the rules and weights together from data. 
  Relational reasoning layers (in the deep learning framework), on the other hand, lose expressiveness as it is hard to comprehend what rules are being learnt. An important need for building real-world AI applications, is to support counterfactual and causal queries. Reasoning mechanisms such as MLN and ProbLog can take cues from P-log (lite) to accommodate such reasoning. %do-calculus.
  
   For human beings, image understanding is a cognitive process that identifies concepts from previous encounters or knowledge. The process goes beyond data-driven pattern matching processes, and delves into the long-standing quest on integrating bottom-up signal processing with top-down knowledge retrieval and reasoning. 
    In this work, we discussed various types of reasoning mechanisms  used by researchers in computer vision to aid a variety of image understanding tasks, ranging from segmentation to QA. %We provide a summary of the discussed applications, corresponding knowledge types and reasoning mechanisms in Table \ref{survey-summary}. 
    %Even though, the utilities and benefits of external knowledge is often acknowledged by several groups of researchers, 
    To conclude, we suggest the following further research pathways to address the observed limitations:
    %list the following observed limitations in the current literature, and thus suggest further research along the pathways to address them: 
    i) speeding up of inference (in MLN, ProbLog, etc. and integrating rule-learning (such as in ILP) will accelerate adoption in vision, % has been scarce, limiting the possibility of performing complex reasoning tasks, 
    ii) scalable reasoning on large common-sense knowledge graphs; %has been limited to using some specific subset of relations for specific applications; 
    iii) probabilistic logical mechanisms supporting counterfactual,  causal and arithmetic queries, enhancing possibilities for higher-level reasoning on real-world datasets. %for higher-level applications such as captioning and QA, only end-to-end architectures have been prominent as state-of-the-art mechanisms and they often suffer from the lack of interpretability and lack of modeling of external knowledge. 
    
    %With the backdrop of this survey, in the rest of the thesis, we describe the knowledge and reasoning mechanisms adopted by our approaches and demonstrate how our approaches perform in large-scale public state-of-the-art datasets.
    
\section*{Acknowledgements}
The support of the National Science Foundation under the Robust Intelligence Program (1816039 and 1750082), research gifts from Verisk AI, and support from Adobe Research (for the first author) are gratefully acknowledged.

\bibliographystyle{plain}
\bibliography{dis}

\begin{thebibliography}{10}

\bibitem{aditya2018spatial}
Somak Aditya, Rudra Saha, Yezhou Yang, and Chitta Baral.
\newblock Spatial knowledge distillation to aid visual reasoning.
\newblock {\em IEEE Winter Conference on Applications of Computer Vision
  (WACV)}, pages 227--235, 2019.

\bibitem{aditya2018vqa}
Somak Aditya, Yezhou Yang, and Chitta Baral.
\newblock {Explicit Reasoning over End-to-End Neural Architectures for Visual
  Question Answering}.
\newblock In {\em AAAI}, pages 629--637, 2018.

\bibitem{aditya2018answering}
Somak Aditya, Yezhou Yang, Chitta Baral, and Yiannis Aloimonos.
\newblock Combining knowledge and reasoning through probabilistic soft logic
  for image puzzle solving.
\newblock In {\em UAI 2018}, pages 238--248. Association For Uncertainty in
  Artificial Intelligence (AUAI), 2018.

\bibitem{aditya2017image}
Somak Aditya, Yezhou Yang, Chitta Baral, Yiannis Aloimonos, and Cornelia
  Fermüller.
\newblock Image understanding using vision and reasoning through scene
  description graph.
\newblock {\em Computer Vision and Image Understanding}, pages 33--45, 2017.

\bibitem{baader2003description}
Franz Baader, Diego Calvanese, Deborah McGuinness, Peter Patel-Schneider, and
  Daniele Nardi.
\newblock {\em The description logic handbook: Theory, implementation and
  applications}.
\newblock Cambridge university press, 2003.

\bibitem{bach2017hinge}
Stephen~H Bach, Matthias Broecheler, Bert Huang, and Lise Getoor.
\newblock Hinge-loss markov random fields and probabilistic soft logic.
\newblock {\em Journal of Machine Learning Research}, 18:1--67, 2017.

\bibitem{Cadene_2019_CVPR}
Remi Cadene, Hedi Ben-Younes, Nicolas Thome, and Matthieu Cord.
\newblock Murel: {M}ultimodal {R}elational {R}easoning for {V}isual {Q}uestion
  {A}nswering.
\newblock In {\em {IEEE} Conference on Computer Vision and Pattern Recognition
  {CVPR}}, 2019.

\bibitem{dasiopoulou2009applying}
Stamatia Dasiopoulou, Ioannis Kompatsiaris, and Michael~G Strintzis.
\newblock Applying fuzzy dls in the extraction of image semantics.
\newblock In {\em Journal on Data Semantics XIV}, pages 105--132. Springer,
  2009.

\bibitem{Davis:2015:CRC:2817191.2701413}
Ernest Davis and Gary Marcus.
\newblock Commonsense reasoning and commonsense knowledge in artificial
  intelligence.
\newblock {\em Commun. ACM}, 58(9):92--103, August 2015.

\bibitem{de2015applying}
Maaike de~Boer, Laura Daniele, Paul Brandt, and Maya Sappelli.
\newblock Applying semantic reasoning in image retrieval.
\newblock {\em Proc. ALLDATA}, 2015.

\bibitem{DeRaedt:2007:PPP:1625275.1625673}
Luc De~Raedt, Angelika Kimmig, and Hannu Toivonen.
\newblock Problog: A probabilistic prolog and its application in link
  discovery.
\newblock In {\em Proceedings of the 20th International Joint Conference on
  Artifical Intelligence}, IJCAI'07, pages 2468--2473, San Francisco, CA, USA,
  2007. Morgan Kaufmann Publishers Inc.

\bibitem{gupta2009observing}
Abhinav Gupta, Aniruddha Kembhavi, and Larry~S Davis.
\newblock Observing human-object interactions: Using spatial and functional
  compatibility for recognition.
\newblock {\em IEEE Transactions on Pattern Analysis and Machine Intelligence},
  31(10):1775--1789, 2009.

\bibitem{havasi2007conceptnet}
Catherine Havasi, Robert Speer, and Jason Alonso.
\newblock Conceptnet 3: a flexible, multilingual semantic network for common
  sense knowledge.
\newblock In {\em Recent advances in natural language processing}, pages
  27--29. Citeseer, 2007.

\bibitem{hinton14071698}
Geoffrey Hinton, Oriol Vinyals, and Jeff Dean.
\newblock Distilling the knowledge in a neural network.
\newblock {\em stat}, 1050:9, 2015.

\bibitem{hudson2018gqa}
Drew~A Hudson and Christopher~D Manning.
\newblock Gqa: A new dataset for real-world visual reasoning and compositional
  question answering.
\newblock {\em Conference on Computer Vision and Pattern Recognition (CVPR)},
  2019.

\bibitem{johnson2017clevr}
Justin Johnson, Bharath Hariharan, Laurens van~der Maaten, Li~Fei-Fei,
  C~Lawrence~Zitnick, and Ross Girshick.
\newblock Clevr: A diagnostic dataset for compositional language and elementary
  visual reasoning.
\newblock In {\em Proceedings of the IEEE Conference on Computer Vision and
  Pattern Recognition}, pages 2901--2910, 2017.

\bibitem{johnsoncvpr2015}
Justin Johnson, Ranjay Krishna, Michael Stark, Jia Li, Michael Bernstein, and
  Li~Fei-Fei.
\newblock Image retrieval using scene graphs.
\newblock In {\em IEEE Conference on Computer Vision and Pattern Recognition
  (CVPR)}, pages 3668--3678, June 2015.

\bibitem{krishnavisualgenome}
Ranjay Krishna, Yuke Zhu, Oliver Groth, Justin Johnson, Kenji Hata, Joshua
  Kravitz, Stephanie Chen, Yannis Kalantidis, Li-Jia Li, David~A. Shamma,
  Michael~S. Bernstein, and Li~Fei-Fei.
\newblock Visual genome: Connecting language and vision using crowdsourced
  dense image annotations.
\newblock {\em Int. J. Comput. Vision}, 123(1):32--73, May 2017.

\bibitem{le2013exploiting}
Dieu-Thu Le, Jasper Uijlings, and Raffaella Bernardi.
\newblock Exploiting language models for visual recognition.
\newblock In {\em Proceedings of the 2013 Conference on Empirical Methods in
  Natural Language Processing}, pages 769--779, 2013.

\bibitem{lee2017computing}
Joohyung Lee, Samidh Talsania, and Yi~Wang.
\newblock Computing lp mln using asp and mln solvers.
\newblock {\em Theory and Practice of Logic Programming}, 17(5-6):942--960,
  2017.

\bibitem{Lenat:1995:CLI:219717.219745}
Douglas~B. Lenat.
\newblock Cyc: A large-scale investment in knowledge infrastructure.
\newblock {\em Commun. ACM}, 38(11):33--38, November 1995.

\bibitem{london2013collective}
Ben London, Sameh Khamis, Stephen Bach, Bert Huang, Lise Getoor, and Larry
  Davis.
\newblock Collective activity detection using hinge-loss markov random fields.
\newblock In {\em Proceedings of the IEEE CVPR Workshops}, pages 566--571,
  2013.

\bibitem{manhaeve2018deepproblog}
Robin Manhaeve, Sebastijan Dumancic, Angelika Kimmig, Thomas Demeester, and Luc
  De~Raedt.
\newblock Deepproblog: Neural probabilistic logic programming.
\newblock In {\em Advances in Neural Information Processing Systems}, pages
  3753--3763, 2018.

\bibitem{marino2017more}
Kenneth Marino, Ruslan Salakhutdinov, and Abhinav Gupta.
\newblock The more you know: Using knowledge graphs for image classification.
\newblock In {\em Proceedings of the IEEE Conference on Computer Vision and
  Pattern Recognition}, pages 2673--2681, 2017.

\bibitem{Miller:1995:WLD:219717.219748}
George~A. Miller.
\newblock Wordnet: A lexical database for english.
\newblock {\em Commun. ACM}, 38(11):39--41, November 1995.

\bibitem{Randell92aspatial}
David~A. Randell, Zhan Cui, and Anthony~G. Cohn.
\newblock A spatial logic based on regions and connection.
\newblock In {\em Proceedings 3rd International Conference ON Knowledge
  Representation And Reasoning}, 1992.

\bibitem{richardson2006markov}
Matthew Richardson and Pedro Domingos.
\newblock Markov logic networks.
\newblock {\em Machine learning}, 62(1-2):107--136, 2006.

\bibitem{NIPS2017_6969}
Tim Rockt{\"a}schel and Sebastian Riedel.
\newblock End-to-end differentiable proving.
\newblock In {\em Advances in Neural Information Processing Systems}, pages
  3788--3800, 2017.

\bibitem{shahMYP19}
Naganand~Yadati Sanket~Shah, Anand~Mishra and Partha~Pratim Talukdar.
\newblock Kvqa: Knowledge-aware visual question answering.
\newblock In {\em AAAI}, 2019.

\bibitem{santoro2017simple}
Adam Santoro, David Raposo, David~G Barrett, Mateusz Malinowski, Razvan
  Pascanu, Peter Battaglia, and Timothy Lillicrap.
\newblock A simple neural network module for relational reasoning.
\newblock In {\em {NIPS}}, pages 4967--4976, 2017.

\bibitem{serafini2016logic}
Luciano Serafini and Artur~d'Avila Garcez.
\newblock Logic tensor networks: Deep learning and logical reasoning from data
  and knowledge.
\newblock {\em arXiv preprint arXiv:1606.04422}, 2016.

\bibitem{suchan2016geometry}
Jakob Suchan and Mehul Bhatt.
\newblock The geometry of a scene: On deep semantics for visual perception
  driven cognitive film, studies.
\newblock In {\em 2016 IEEE Winter Conference on Applications of Computer
  Vision (WACV)}, pages 1--9. IEEE, 2016.

\bibitem{Suchanek:2007:YCS:1242572.1242667}
Fabian~M. Suchanek, Gjergji Kasneci, and Gerhard Weikum.
\newblock Yago: A core of semantic knowledge.
\newblock In {\em Proceedings of the 16th International Conference on World
  Wide Web}, WWW '07, pages 697--706, New York, NY, USA, 2007. ACM.

\bibitem{summers2012using}
Douglas Summers-Stay, Ching~L Teo, Yezhou Yang, Cornelia Ferm{\"u}ller, and
  Yiannis Aloimonos.
\newblock Using a minimal action grammar for activity understanding in the real
  world.
\newblock In {\em 2012 IEEE/RSJ International Conference on Intelligent Robots
  and Systems}, pages 4104--4111. IEEE, 2012.

\bibitem{wang2017fvqa}
Peng Wang, Qi~Wu, Chunhua Shen, Anthony Dick, and Anton van~den Hengel.
\newblock Fvqa: fact-based visual question answering.
\newblock {\em IEEE TPAMI}, 2017.

\bibitem{wu2017visual}
Qi~Wu, Damien Teney, Peng Wang, Chunhua Shen, Anthony Dick, and Anton van~den
  Hengel.
\newblock Visual question answering: A survey of methods and datasets.
\newblock {\em Computer Vision and Image Understanding}, 163:21--40, 2017.

\bibitem{wu2016ask}
Qi~Wu, Peng Wang, Chunhua Shen, Anthony Dick, and Anton van~den Hengel.
\newblock Ask me anything: Free-form visual question answering based on
  knowledge from external sources.
\newblock In {\em IEEE Conference on Computer Vision and Pattern Recognition
  {CVPR}}, pages 4622--4630, 2016.

\bibitem{wulfmeier2016incorporating}
Markus Wulfmeier, Dushyant Rao, and Ingmar Posner.
\newblock Incorporating human domain knowledge into large scale cost function
  learning.
\newblock {\em arXiv preprint arXiv:1612.04318}, 2016.

\bibitem{yu2017_vrd_knowledge_distillation}
Ruichi Yu, Ang Li, Vlad~I. Morariu, and Larry~S. Davis.
\newblock {Visual Relationship Detection with Internal and External Linguistic
  Knowledge Distillation.}
\newblock {\em ICCV}, 2017.

\bibitem{zheng2015scene}
Bo~Zheng, Yibiao Zhao, Joey Yu, Katsushi Ikeuchi, and Song-Chun Zhu.
\newblock Scene understanding by reasoning stability and safety.
\newblock {\em International Journal of Computer Vision}, 112(2):221--238,
  2015.

\bibitem{Zhou_2018_ECCV}
Bolei Zhou, Alex Andonian, Aude Oliva, and Antonio Torralba.
\newblock Temporal relational reasoning in videos.
\newblock In {\em ECCV}, September 2018.

\bibitem{conf/eccv/ZhuFF14}
Yuke Zhu, Alireza Fathi, and Li~Fei-Fei.
\newblock Reasoning about object affordances in a knowledge base
  representation.
\newblock In {\em ECCV (2)}, pages 408--424. Springer, 2014.

\end{thebibliography}

\end{document}